\documentclass[final]{cvpr}

\usepackage{times}
\usepackage{epsfig}
\usepackage{graphicx}
\usepackage{amsmath}
\usepackage{amssymb}

\usepackage[pagebackref=true,breaklinks=true,colorlinks,bookmarks=false]{hyperref}

\def\cvprPaperID{27} 
\def\confYear{CVPR 2021}
\begin{document}

\title{Stacked Deep Multi-Scale Hierarchical Network for Fast Bokeh Effect Rendering from a Single Image}

\author{Saikat Dutta\\
IIT Madras\\
Chennai, India\\
{\tt\small saikat.dutta779@gmail.com}
\and
Sourya Dipta Das\\
Jadavpur University\\
Kolkata, India\\
{\tt\small dipta.juetce@gmail.com}
\and
Nisarg A. Shah\\
IIT Jodhpur\\
Jodhpur, India\\
{\tt\small shah.2@iitj.ac.in}
\and
Anil Kumar Tiwari\\
IIT Jodhpur\\
Jodhpur, India\\
{\tt\small akt@iitj.ac.in}
}

\maketitle

\begin{abstract}
The Bokeh Effect is one of the most desirable effects in photography for rendering artistic and aesthetic photos. Usually, it requires a DSLR camera with different aperture and shutter settings and certain photography skills to generate this effect. In smartphones, computational methods and additional sensors are used to overcome the physical lens and sensor limitations to achieve such effect. Most of the existing methods utilized additional sensor’s data or pretrained network for fine depth estimation of the scene and sometimes use portrait segmentation pretrained network module to segment salient objects in the image. Because of these reasons, networks have many parameters, become runtime intensive and unable to run in mid-range devices. In this paper, we used an end-to-end  Deep Multi-Scale Hierarchical Network (DMSHN) model for direct Bokeh effect rendering of images captured from the monocular camera. To further improve the perceptual quality of such effect, a stacked model consisting of two DMSHN modules is also proposed. Our model does not rely on any pretrained network module for Monocular Depth Estimation or Saliency Detection, thus significantly reducing the size of model and run time. Stacked DMSHN achieves state-of-the-art results on a large scale EBB! dataset with around 6x less runtime compared to the current state-of-the-art model in processing HD quality images.
\end{abstract}

\section{Introduction}
The word ``Bokeh" originated from the Japanese word ``boke" which means blur. In photography, Bokeh effect refers to the pleasing or aesthetic quality of the blur produced in the out-of-focus parts of an image produced by a camera lens. These images are usually captured with DSLR cameras using a large focal length and a large aperture size. Unlike DSLR cameras, mobile phone cameras have limitations on its size and weight. Thus, mobile phones cannot generate the same quality of bokeh as DSLR can do as the mobile cameras have a small and fixed-size aperture that produces images with the scene mostly in its focus, depending on depth map of the scene.
One approach to solve these limitations is to computationally simulate the bokeh effect on the mobile devices with small camera. There are some works where additional depth information obtained from a dual pixel camera \cite{wadhwa2018synthetic} or stereo camera (one main camera and one calibrated subcamera) \cite{luo2020wavelet,busam2019sterefo,liu2015bokeh} is incorporated in the model to simulate a more realistic bokeh image. However, these systems suffer from a variety of disadvantages like (a) addition of a stereo camera or depth sensor increases the size and the cost of the device and power consumption during usage (b) structured light depth sensors can be affected by poor resolution or high sensitivity to interference with ambient light (c) it performs badly when the target of interest is substantially distant from the camera, it is difficult to estimate good depth map for further regions leading because of the small stereo baseline of stereo camera (d) these approaches can't be used to enhance pictures already taken with monocular cameras as post processing since depth information is often not saved along with the picture.

To address these problems, we propose Deep Multi-Scale Hierarchical Network (DMSHN) for Bokeh effect rendering from the monocular lens without using any specialized hardware. Here, the proposed model synthesizes bokeh effect under the ``coarse-to-fine” scheme by exploiting multi-scale input images at different processing levels. Each lower level acts in the residual manner by contributing its
residual image to the higher level thus with intermediate feature aggregation. In this way, low level encoders can provide additional global intermediate features to higher level for improving saliency, and higher level encoders can provide more local intermediate features to improve fidelity of generated bokeh images.
Our model does not depend on any Monocular Depth Estimation or Saliency Detection pretrained network module, hence reducing the number of parameters and runtime considerably. We have also explored a stacked version of DMSHN, namely Stacked DMSHN, where  two DMSHN modules were connected horizontally to boost the performance. It achieves state-of-the-art results in Bokeh Effect Rendering on monocular images.
Though this improvement in accuracy in stacked DMSHN comes with increased runtime and model parameters, still DMSHN and Stacked DMSHN can process HD quality images in 50 fps and 25 fps, respectively, which is significantly faster than other methods in the literature. We have also shown the runtime of our models deployed in mid-range smartphone devices to demonstrate its efficiency. In our experiments, we have used a large-scale \textit{EBB!} dataset \cite{pynet} containing more than 10,000 images collected in the wild with the DSLR camera.

\section{Related Work}
 Many works in Bokeh Effect Rendering leveraged depth information from images captured by two cameras. Liu \emph{et al.} \cite{liu2015bokeh} presented a bokeh simulation method by using depth estimation map through stereo matching. They have also designed a convenient bokeh control interface that uses a little user interaction for identifying the salient object and control the bokeh effect by setting a specific kernel. Busam \emph{et al.} \cite{busam2019sterefo} also proposed a stereo vision-based fast and efficient algorithm to perform the refocusing by using high-quality disparity map via efficient stereo depth estimation. Recently, Luo \emph{et al.} \cite{luo2020wavelet} proposed a novel deep neural architecture named wavelet synthesis neural network (WSN), to produce high-quality disparity maps on smartphones by using a pair of calibrated stereo images. However, these methods are ineffective in the case of monocular cameras or in post-processing of previously captured images.
 \begin{figure*}[h]
    \centering
    \includegraphics[width=0.8\textwidth]{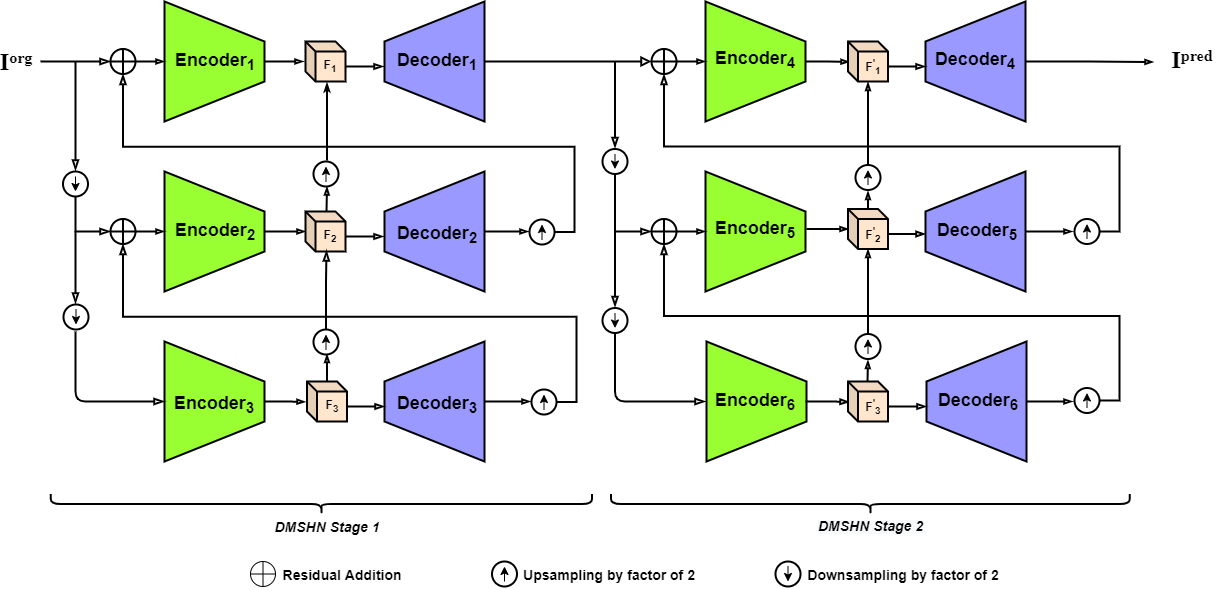}
    \caption{Stacked Deep Multi-Scale Hierarchical Network.}
    \label{dmshn}
\end{figure*}

In one of the earliest works in Monocular Bokeh Effect Rendering, Shen \emph{et al.} \cite{shen2016automatic} used a Fully Convolutional Network to perform portrait image segmentation and using this segmentation map, they generated depth-of-field image with uniformly blurred background on portrait images. Wadhwa \emph{et al.} \cite{wadhwa2018synthetic} proposed a system to generate synthetic shallow depth-of-field images on mobile devices by incorporating a portrait segmentation network and depth map from the camera's dual-pixel auto-focus system. But their system's limitation is that their bokeh rendering method is not photorealistic as actual bokeh photos taken from DSLR cameras because of using an approximated disk blur kernel to blur the background. Xu \emph{et al.} \cite{xu2018rendering} also proposed a similar approach where they had used two different deep neural networks to get depth map and portrait segmentation map from a single image. Then, they improved those initial estimates using another deep neural network and trained a depth and segmentation guided Recursive Neural Network to approximate and accelerate the bokeh rendering. 

Lijun \emph{et al.} \cite{lijun2018deeplens} presented a memory efficient deep neural network architecture with a lens blur module, which synthesized the lens blur and guided upsampling module to generate bokeh images at high resolution with user controllable camera parameters at interactive speed of rendering. 
Dutta \cite{dutta2020depth}  formulated bokeh image as a weighted sum of the input image and its different blurred versions obtained by using different sizes of Gaussian blur kernels. The weights for this purpose were generated using a fine-tuned depth estimation network. Purohit \emph{et al.} \cite{purohit2019depth} designed a model consisting of a densely connected encoder and decoder taking benefits of joint Dynamic Filtering and intensity estimation for the spatially-aware background blurring. Their model utilized pretrained networks for depth estimation and saliency map segmentation to guide the bokeh rendering process.
Zheng \emph{et al.} \cite{ignatov2019aim} used a multi-scale predictive filter CNN consisting of Gate Fusion Block, Constrained Predictive Filter Block, and Image Reconstruction Block. They trained their model using image patches with concatenated pixel coordinate maps with respect to the full-scale image. Xiong \emph{et al.} \cite{ignatov2019aim} used an ensemble of modified U-Net consisting of residual attention mechanism, multiple Atrous Spatial Pyramid Pooling blocks, and Fusion Modules. Yang \emph{et al.} \cite{ignatov2019aim} used two stacked bokehNet with additional memory blocks to capture global and local features. The generated feature maps from each model are concatenated and fed to a Selective Kernel Network \cite{li2019selective}.

Ignatov \emph{et al.} \cite{pynet} presented a multi-scale end-to-end deep learning architecture, PyNet, for natural bokeh image rendering by utilizing both input image captured with narrow aperture and pre-computed depth map of the same image. In this paper, we propose a fast, lightweight efficient network, Stacked DMSHN for Single-image Bokeh effect rendering. Our model doesn't require additional depth maps which makes it suitable for post-processing photos.

\begin{figure*}
    \centering
    \includegraphics[width=0.8\textwidth]{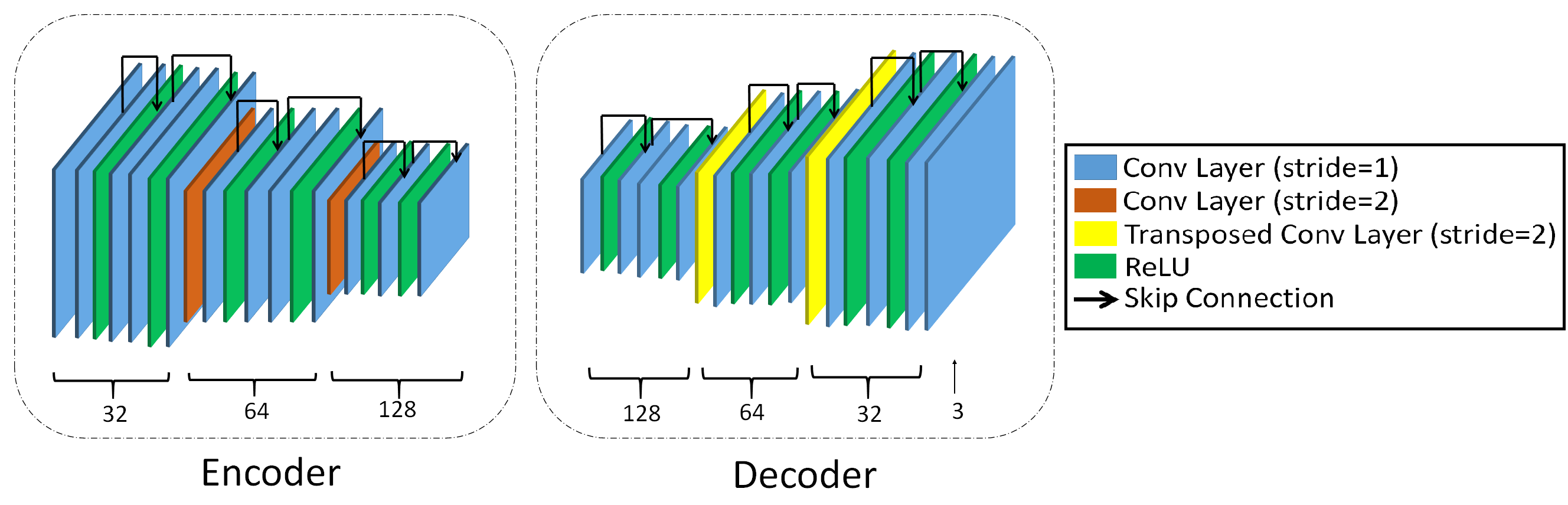}
    \caption{Encoder and Decoder Architecture. Numbers below the curly braces denote number of output channels of the corresponding layers.}
    \label{enc_dec}
\end{figure*}

\section{Proposed Method}
We used Stacked Deep Multi-Scale Hierarchical Network (DMSHN) for Bokeh Effect Rendering. The model diagram of Stacked DMSHN is shown in Fig. \ref{dmshn}. Stacked DMSHN consists of two DMSHN base networks which are cascaded one after another. The details of DMSHN architecture are described in the following.

\textbf{Base Network:} We used Deep Multi-Scale Hierarchical Network (DMSHN) as the base network. The network works on a 3-level Image pyramid. Let the input images at $i^{th}$ level be denoted as $I^{org}_i$.
\begin{align}
    I^{org}_1 = I^{input}\\
    I^{org}_{2} = down(I^{org}_{1})\\
    I^{org}_{3} = down(I^{org}_{2})
\end{align}
where $down(.)$ is bilinear downsampling by factor of two.

In each level of the network, we have an encoder and a decoder. Let's denote these modules as $Encoder_i$ and $Decoder_i$ respectively for $i^{th}$ level.  

At the bottom-most level encoder, $Encoder_3$, takes bokeh-free image $I^{org}_3$ as input. The features $F_3$ generated by $Encoder_3$ is then fed into $Decoder_3$. 
So, the generated feature map $F_3$ is then fed to $Decoder_3$ to generate residual features $res_3$.
\begin{equation}
    F_3=Encoder_3(I^{org}_3)
\end{equation}
\begin{equation}
    res_3 = Decoder_3(F_3)
\end{equation}
$res_3$ is upsampled by a factor of $2$ and then added to $I^{org}_2$ in the next level and fed to $Encoder_2$. The encoded feature map $G_2$ is then added to upscaled feature $F_3$ to produce $F_2$. Reusing the encoded features from the lower level helps the network leverage the global context information learned by the encoder in the previous level. Residual connection between encoders of different levels is useful in correct localization and reconstruction of the foreground. $F_2$ is then passed to $Decoder_2$ which generates residual features $res_2$.
\begin{equation}
    G_2 = Encoder_2(I^{org}_2 + up(res_3))
\end{equation}
\begin{equation}
    F_2 = G_2 + up(F_3) 
\end{equation}
\begin{equation}    
    res_2 = Decoder_2(F_2)
\end{equation}

where $up(.)$ is bilinear upsampling by factor of $2$.
The upscaled residual feature map $res_2$ is added to $I^{org}_1$ in the top-most level and passed to $Encoder_1$ to generate encoded feature map $G_1$. $G_1$ is added with upscaled $F_2$ to generate $F_1$. $F_1$ is further fed to $Decoder_1$ to generate the final bokeh image $I^{pred}$.

\begin{equation}
    G_1 = Encoder_1(I^{org}_1 + up(res_2))
\end{equation}
\begin{equation}
F_1 = G_1 + up(F_2)
\end{equation}
\begin{equation}
I^{pred} = Decoder_1(F_1)
\end{equation}
For the task of non-homogeneous Image Dehazing \cite{ancuti2020ntire}, DMSHN \cite{dipta2020fast} was coined for ablation studies. In \cite{dipta2020fast}, the authors have shown the inferiority of DMSHN with respect to a Patch-hierarchical network. However we have shown in this paper that this type of multi-scale architecture is more suitable for solving problems where capturing the global context and saliency in the image is important, by incorporating it to Bokeh Effect Rendering.

\textbf{Encoder and Decoder Architecture: }
The encoder and the decoder consists of 3 level and each level consists of two Residual blocks. Convolutional layers with stride 2 is used to decrease spatial resolution of the feature maps in the encoder and Transposed convolution is used for increasing feature map resolution in the decoder. ReLU is used as activation function and kernel size used in convolutional layers is $3 \times 3$ everywhere. Same architecture of encoder and decoder is used at all the levels of our network. The architecture diagram of encoder and decoder is shown in Fig. \ref{enc_dec}. 

Instead of increasing depth of DMSHN model vertically, stacking two or more DMSHN module horizontally can significantly improve the performance \cite{zhang2019deep}. Cascading multiple base networks can help refining the rendered bokeh images. We cascaded two pretrained DMSHN models in this approach and finetune the whole network. Let the first network be denoted as $net_1$ and the second one as $net_2$. The final output $I^{pred}$ is given by, 
\begin{equation}
I^{pred} = net_2(net_1(I^{org}_1))
\end{equation}

Our Stacked DMSHN model has a few similarities with Gridnet \cite{gridnet}. However the key differences are: (a) Feature maps at the same level of Gridnet has
same channel and spatial dimension, which is not the
case for Stacked DMSHN (b) 2nd and 3rd levels of DMSHN blocks are not connected in Stacked
DMSHN.

\section{Experiments}
\subsection{System description}
We implemented the proposed models in Python and PyTorch \cite{pytorch}. Our models were trained on a machine with Intel Xeon CPU with 16 GB RAM and NVIDIA 1080Ti GPU card with approximately 12 GB GPU Memory. 
\subsection{Dataset Description}
Everything is Better with Bokeh (\textit{EBB!}) Dataset \cite{pynet} is used in this work. In this dataset, the images were taken from a wide range of locations during the daytime, and in varying lighting and weather conditions. This dataset contains 5094 pairs of Bokeh-free and Bokeh images. The training set consists of 4694 image pairs whereas the validation and test set contains 200 image pairs each. The Bokeh-free images were captured using a narrow aperture (f/16) and the corresponding bokeh images were captured using high aperture (f/1.8). The average image resolution of this dataset is $1024 \times 1536$. Since the Validation and test set ground truth images are not available yet, we use \textit{Val294} set \cite{dutta2020depth} was used for evaluation and the rest of the dataset is used for training the models.

\subsection{Training and Testing Details}
We rescaled the images to $1024 \times 1024$ for training. Data augmentation techniques e.g., horizontal and vertical flipping were used to increase the training set size. We use Adam optimizer \cite{kingma2014adam} with $\beta_1=0.9$ and $\beta_2=0.999$ for training the network with a batch size of 2. The initial learning rate is set to $10^{-4}$ and gradually decreased to $10^{-6}$. 

During inference time, the input images are resized to $1024 \times 1536$ and fed into the network. The output from the network is rescaled back to its original dimension using bilinear interpolation.

\subsection{Loss functions}
The proposed network was trained in two stages. We used a linear combination of L1 loss and SSIM loss \cite{zhao2016loss} in the first stage. L1 loss helps in pixelwise reconstruction of the synthesized bokeh image. SSIM loss is used to improve perceptual quality of the generated image, because it focuses on the similarity of local structure. So, the loss function used in the first stage is given by, 
\begin{equation}
\mathcal{L}_{st1} = \mathcal{L}_1 + \alpha.\mathcal{L}_{SSIM}
\end{equation}
where $ \mathcal{L}_1 = ||I^{pred} - I^{gt}||_1 $, $ \mathcal{L}_{SSIM} = 1 - SSIM(I^{pred} , I^{gt}) $ and $I^{gt}$ is the ground truth bokeh image.
In the second stage, we use Multi-scale SSIM (MS-SSIM) loss \cite{zhao2016loss} with default scale value of $5$, to further fine tune the model. MS-SSIM loss is based on MS-SSIM \cite{msssim} which considers structural similarity at multiple levels of image pyramid. So, the loss in second stage is given by, 
\begin{equation} \label{eq1}
\begin{split}
\mathcal{L}_{st2} & = \mathcal{L}_{MS\text{\small-}SSIM} \\
 & = 1 - MS\text{\small-}SSIM(I^{pred} , I^{gt}) 
\end{split}
\end{equation}
In our experiments, $\alpha$ is chosen to be $0.1$. 

For the training of Stacked DMSHN network, we loaded the weights of pretrained DMSHN model and finetuned the whole network using $\mathcal{L}_{MS\text{\small-}SSIM}$.


\begin{figure*}[!h!t!p]
    \centering
    \includegraphics[width=0.95\textwidth]{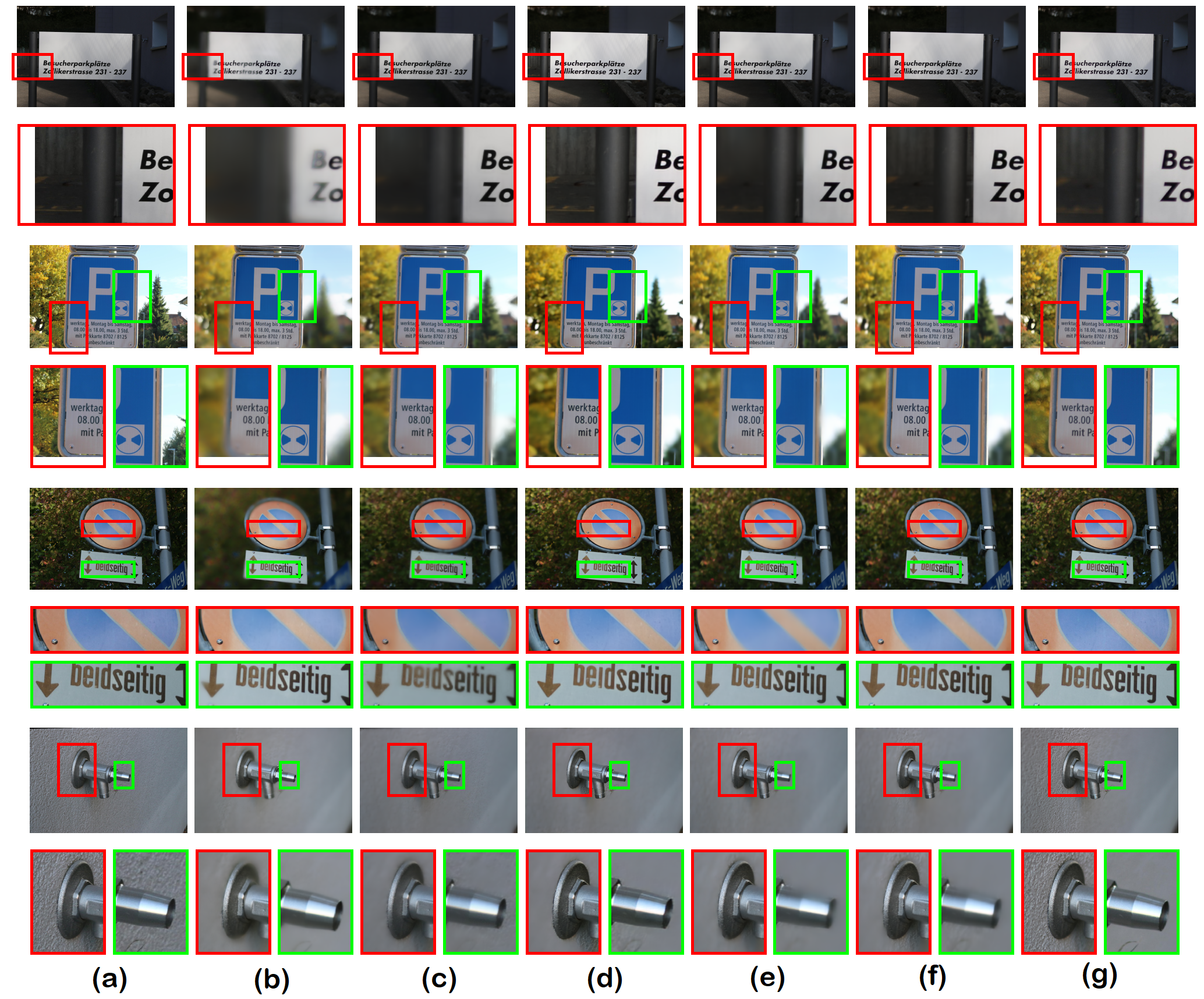}
    \caption{Comparison with other methods. From left: (a) Input Image (b) SKN \cite{ignatov2019aim} (c) DBSI \cite{dutta2020depth} (d) PyNet \cite{pynet} (e) DMSHN (ours) (f) Stacked DMSHN (ours) (g) Ground Truth.}
    \label{sota_comp}
\end{figure*}

\subsection{Results}
\subsubsection{Evaluation metrics:} We used Peak signal-to-noise ratio (PSNR), Structural Similarity (SSIM) \cite{ssim_paper} and Learned Perceptual Image Patch Similarity metrics (LPIPS) \cite{LPIPS} for comparative evaluation.
For higher similarity with respect to ground truth, higher values of PSNR and SSIM, and lower values of LPIPS scores are desired.
Although LPIPS is often used as Perceptual Metric in Image Restoration problems, this is not a reliable metric in case of Bokeh Effect Rendering as discussed in  \cite{pynet}. Thus, we also evaluated our models on Mean Opinion Score (MOS) based on a user study. 

\textbf{User Study:} 25 users having good experience in photography were asked to rate images with bokeh effect generated from different methods. The users were presented with 20 sets of images from \textit{Val294} where each set contained bokeh ground truth image and rendered bokeh images. Scoring was done on a scale of 0 to 4 where 0 stands for ``almost similar" and 4 is ``mostly different" (following  \cite{pynet}). The users were suggested to give a low rating (high quality) if (a) All edges, text present in the salient object region are intact in the generated bokeh image (b) There is no particular artifacts (bright spot, blur circle) present in the generated bokeh image.
(c) There are no changes in chromatic feature or brightness of the image.
(d) The salient region from the generated bokeh image is the same as the ground truth image.

\subsubsection{Comparison with other methods} We compared the performance of our method with state-of-the-art methods, e.g. PyNet \cite{pynet}, Depth-aware Blending of Smoothed Images (DBSI) \cite{dutta2020depth} and Selective Kernel networks (SKN) \cite{li2019selective}. As the source code is not available for Depth-guided Dense Dynamic Filtering network (DDDF) \cite{purohit2019depth}, we only show comparison with other methods\footnote{For qualitative comparison on \textit{EBB!} Test data, please refer to Supplementary material.}. Table \ref{tab_sota_comp} shows that our Stacked DMSHN model performs better than DMSHN. Both DMSHN and Stacked DMSHN performs better than SKN \cite{ignatov2019aim} and DBSI \cite{dutta2020depth}, whereas Stacked DMSHN achieves similar perceptual quality to that of PyNet \cite{pynet}. Qualitative comparison (shown in Fig. \ref{sota_comp}) reveals that the Stacked DMSHN is at par with PyNet and better than other methods in sharp reconstruction of the foreground.

\begin{table}[h]
\centering
\small
\begin{tabular}{|c|c|c|c|c|}
\hline
                                                                Method & PSNR         & SSIM           & LPIPS           & MOS \\ \hline
SKN \cite{ignatov2019aim}                                        &  24.66              &    0.8521             &      0.3323           &   3.71  \\ \hline
DBSI \cite{dutta2020depth}                                       & 23.45          & 0.8675          & 0.2463          &  1.89   \\ \hline
PyNet \cite{pynet}                                               & \textcolor{red}{24.93} & \textcolor{blue}{0.8788}          & \textcolor{red}{0.2219} &   \textcolor{red}{1.11}  \\ \hline
DMSHN (ours)                                                    & 24.65          & 0.8765          & 0.2289          &   1.52  \\ \hline
\begin{tabular}[c]{@{}c@{}}Stacked DMSHN \\ (ours)\end{tabular} & \textcolor{blue}{24.72}          & \textcolor{red}{0.8793} & \textcolor{blue}{0.2271}          &   \textcolor{blue}{1.17}  \\ \hline
\end{tabular}
\caption{Quantitative Comparison with other methods on Val294 set. The scores in Red and Blue represent best and second best scores respectively.}
\label{tab_sota_comp}
\end{table}

\begin{figure*}[h] 
    \centering
    \includegraphics[width=0.7\textwidth]{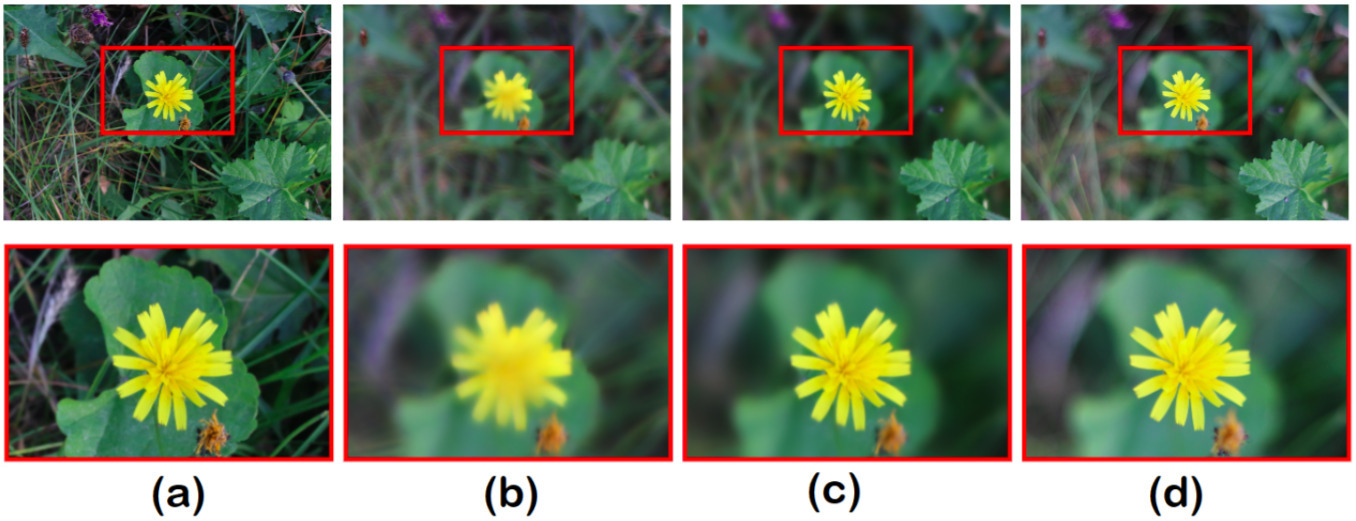}
    \caption{Effect of residual connection between encoders of different levels. From left: (a) Input Image (b) DMSHN (w/o res.) (c) DMSHN (with res.) (d) Ground Truth.}
    \label{abl_res}
\end{figure*}

\begin{figure*}[h]
    \centering
    \includegraphics[width=0.8\textwidth]{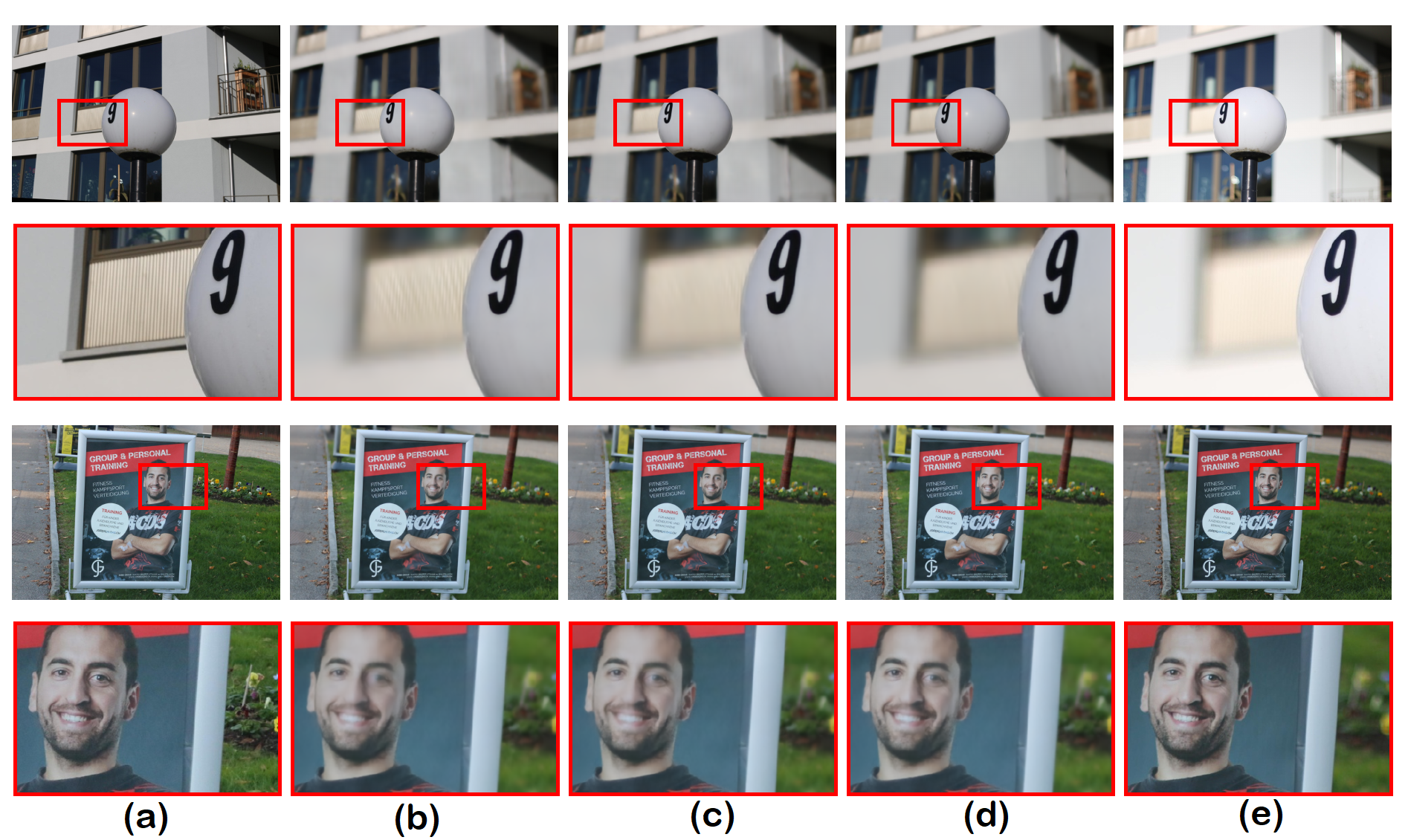}
    \caption{Effect of stage-2 training. From left: (a) Input Image (b) DMSHN (w/o stage-2) (c) DMSHN ($\mathcal{L}_1 + 0.1.\mathcal{L}_{MS\text{\small-}SSIM}$) (d) DMSHN ($\mathcal{L}_{MS\text{\small-}SSIM}$) (e) Ground Truth.}
    \label{loss_abl}
\end{figure*}


\subsection{Ablation study}
\textbf{Importance of residual connections between encoded features:} To show the significance of skip connections between encoded features at different levels, we train one variant of DMSHN where such residual connections are removed. It can be observed from Fig. \ref{abl_res} that residual connections between the encoded features is crucial in correctly detecting the foreground and increasing the overall quality of the rendered bokeh image. Qualitative comparison between the two variants in Table \ref{tab_abl_res} shows significant improvement in performance when the residual connections are used.

\begin{table}[htp]
\centering
\small
\begin{tabular}{|c|c|c|c|c|}
\hline
               Method & PSNR & SSIM & LPIPS & MOS \\ \hline
DMSHN (w/o res.)  &  22.73    &   0.8150   &   0.2953    &  3.34   \\ \hline
DMSHN (with res.) &  \textbf{24.65}    &  \textbf{0.8765}    &   \textbf{0.2289}    &   \textbf{1.52}  \\ \hline
\end{tabular}
\caption{Importance of Residual Connections between encoded features.}
\label{tab_abl_res}
\end{table}

\begin{figure*}[!t]
    \centering
    \includegraphics[width=0.8\textwidth]{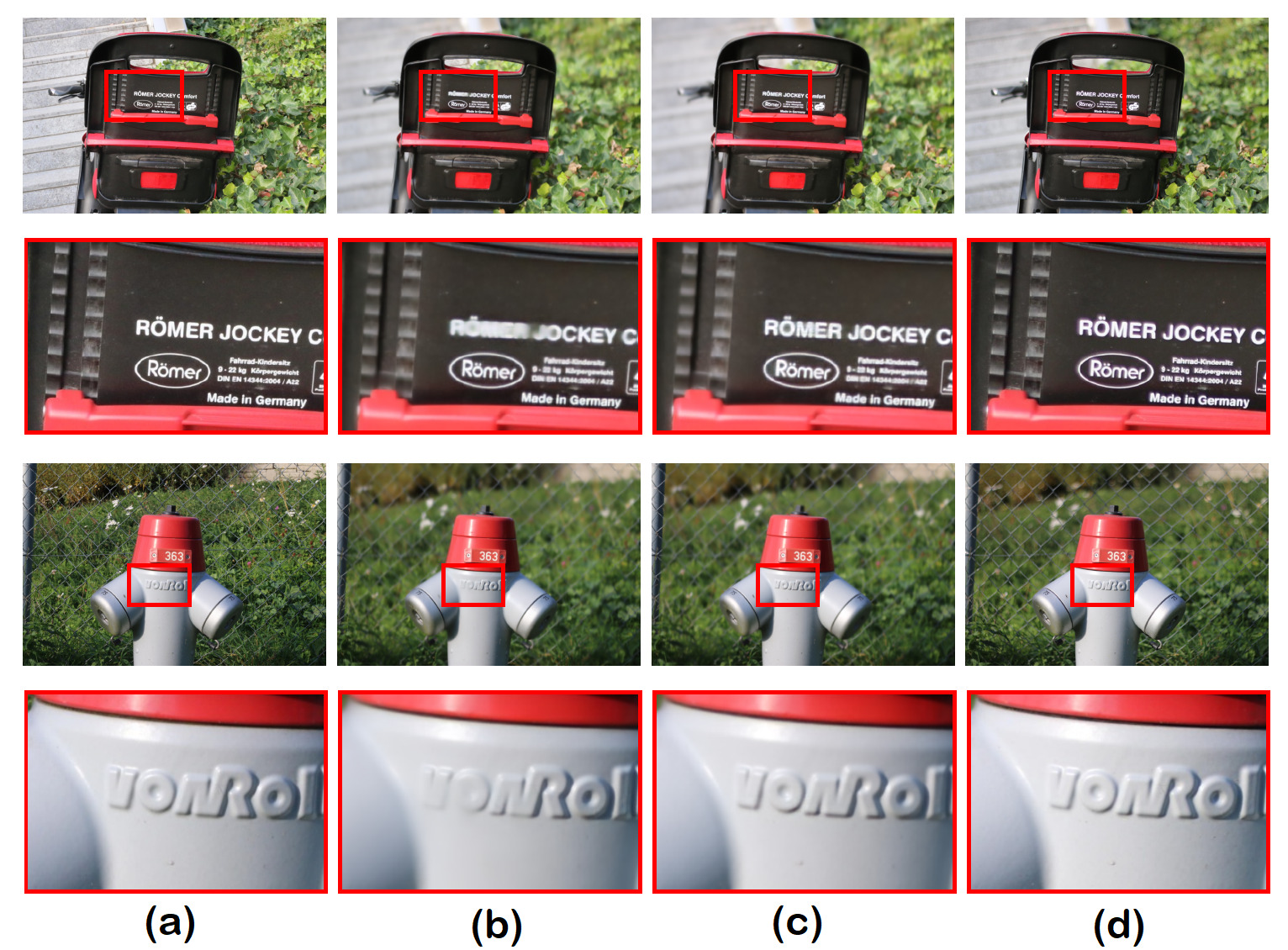}
    \caption{Effect of Stacking. From left: (a) Input Image (b) DMSHN (d) Stacked DMSHN (e) Ground Truth.}
    \label{abl_stack}
\end{figure*}

\textbf{Effect of the second stage of training:} The network is first trained with a combination of $ \mathcal{L}_1 $ and $ \mathcal{L}_{SSIM}$ and then finetuned with $ \mathcal{L}_{MS\text{\small-}SSIM}$. We compare the results of our network with Stage-2
training and without Stage-2 training. We have also experimented with a combination of $\mathcal{L}_1$ and $\mathcal{L}_{MS\text{\small-}SSIM} $ in second stage training. Table \ref{tab_abl_loss} shows that stage-2 training with $\mathcal{L}_1$ and $\mathcal{L}_{MS\text{\small-}SSIM}$ improves SSIM, LPIPS and MOS metrics over no stage-2 training, whereas using $\mathcal{L}_{MS\text{\small-}SSIM}$ alone in stage-2 training improves all the four metrics. From the first and second row of Fig. \ref{loss_abl}, it can be inferred that stage-2 training removes artifacts in the background, and the third and fourth row shows that $\mathcal{L}_{MS\text{\small-}SSIM}$ helps in the better reconstruction of the foreground.

\begin{table}[htp]
\centering
\footnotesize
\renewcommand{\tabcolsep}{3pt}
\renewcommand{\arraystretch}{1.2}
\begin{tabular}{|c|c|c|c|c|}
\hline
                   Method with Specification & PSNR  & SSIM   & LPIPS  & MOS \\ \hline
DMSHN (without stage-2) & 24.41 & \textbf{0.8765} & 0.2322 &  1.71   \\ \hline
\begin{tabular}[c]{@{}c@{}}DMSHN (with stage-2) \\ $\mathcal{L}_{st2} =\mathcal{L}_1  + 0.1*\mathcal{L}_{MS\text{\small-}SSIM}$\end{tabular} & 24.34 & 0.8748 & 0.2299 &  1.67   \\ \hline
\begin{tabular}[c]{@{}c@{}}DMSHN (with stage-2) \\ $\mathcal{L}_{st2} =\mathcal{L}_{MS\text{\small-}SSIM}$\end{tabular} & \textbf{24.65} & \textbf{0.8765} & \textbf{0.2289} &  \textbf{1.52}   \\ \hline
\end{tabular}
\caption{Effect of different Loss functions in Stage-2 training.}
\label{tab_abl_loss}
\end{table}


\textbf{DMSHN vs Stacked DMSHN:} By stacking two DMSHN networks helps recovering important details in the foreground image in the rendered bokeh image as shown in Fig. \ref{abl_stack}. Table \ref{tab_sota_comp} shows the quantitative improvement of Stacked DMSHN over DMSHN.

\textbf{Inclusion of depth maps:} In order to see if depth maps are useful in our networks, normalized depth maps were computed from the input images using a state-of-the-art Monocular Depth estimation network MegaDepth \cite{li2018megadepth}. The estimated depth map was resized and concatenated to encoder input at each scale.

\begin{figure*}[h]
\centering
\includegraphics[width=0.9\textwidth]{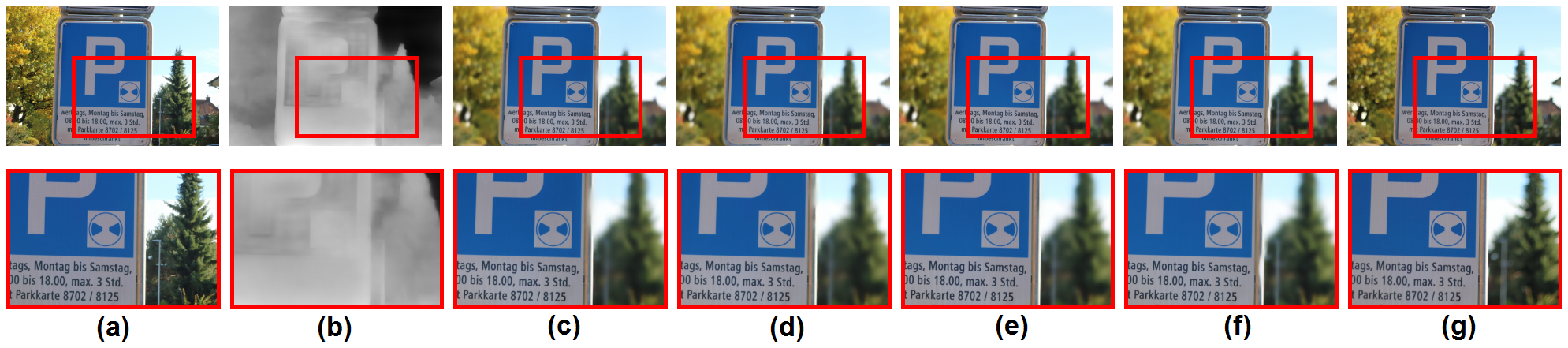}
\caption{Effect of inclusion of depth maps. From left: (a) Input Image (b) Estimated Depth Map (c) DMSHN (d) DMSHN (with depth) (e) Stacked DMSHN (f) Stacked DMSHN (with depth) (g) Ground Truth}
\label{depth_abl_pic}
\end{figure*}

Table \ref{tab_abl_depth} shows that inclusion of depth maps improves the performance of DMSHN. Although LPIPS improves by a small margin in case of Stacked DMSHN after incorporating depth maps, PSNR, SSIM and MOS scores do not improve. It indicates that inaccuracies in depth map estimation doesn't help Stacked DMSHN further in Bokeh rendering. Qualitative comparison in Fig. \ref{depth_abl_pic} shows Stacked DMSHN without depth map produces best perceptual results.

\begin{table}[htp]
\small
\centering
\begin{tabular}{|c|c|c|c|c|}
\hline
Method                                                               & PSNR           & SSIM            & LPIPS           & MOS           \\ \hline
DMSHN                                                                & 24.65          & 0.8765          & 0.2289          & 1.52          \\ \hline
DMSHN (with depth)                                                   & 24.68          & 0.8780          & 0.2264          & 1.39          \\ \hline
Stacked DMSHN                                                        & \textbf{24.72} & \textbf{0.8793} & 0.2271          & \textbf{1.17} \\ \hline
\begin{tabular}[c]{@{}c@{}}Stacked DMSHN\\ (with depth)\end{tabular} & 24.67          & 0.8780          & \textbf{0.2263} & 1.21          \\ \hline
\end{tabular}
\caption{Effect of Inclusion of Depth maps in our network.}
\label{tab_abl_depth}
\end{table}

\subsection{Efficiency and Deployment in Mobile Devices:}
The proposed models, DMSHN and Stacked DMSHN are lightweight as DMSHN has 5.42 million, whereas Stacked DMSHN has 10.84 million trainable parameters. DMSHN and Stacked DMSHN models take 0.02 and 0.04 seconds to process an HD image, respectively on our system. Table \ref{tab_efficiency} shows parameter and runtime comparison with other existing models. DMSHN is faster than all other models. Our final model, Stacked DMSHN is 0.23 seconds faster than the current state-of-the-art PyNet \cite{pynet}. It is also important to note that PyNet takes precomputed depth maps as input, generating which takes additional time if it is not readily available, whereas Stacked DMSHN can process the input image directly.

\begin{table}[h]
\footnotesize
\centering
\begin{tabular}{|c|c|c|}
\hline
                            Method & Parameters (M)         & Runtime (s)                                       \\ \hline
SKN \cite{ignatov2019aim}    & \textcolor{blue}{5.37} & 0.055                                               \\ \hline
DDDF \cite{purohit2019depth} & N/A                    & 2.5 \\ \hline
DBSI \cite{dutta2020depth}   & \textcolor{red}{5.36}  & 0.048                                               \\ \hline
PyNet \cite{pynet}           & 47.5                   & 0.27                                                \\ \hline
DMSHN (ours)                & 5.42                   & \textcolor{red}{0.020}                              \\ \hline
Stacked DMSHN (ours) & 10.84                  & \textcolor{blue}{0.040}                             \\ \hline                             
\end{tabular}
\caption{Parameter and Runtime comparison with state-of-the-art models. Runtime for DDDF is reported from \cite{purohit2019depth} and rest of the runtimes were measured on our system. Red and Blue represent best and second best values respectively.}
\label{tab_efficiency}
\end{table}

Our models are also deployable in mobile devices. We converted our PyTorch models to Tensorflow Lite (TFLite) \cite{abadi2016tensorflow} models and ran them on AI Benchmark Android application \cite{ignatov2018ai}. Here, we have selected three mainstream mid-range mobile chipsets from three different smartphone manufacturers. The configurations of these chipsets are as follows.

\textbf{\textit{Config-1}}: Qualcomm Snapdragon 660 AIE processor, Adreno 512 GPU and 4GB RAM.

\textbf{\textit{Config-2}}: Exynos 9611 processor, Mali-G72 MP3 GPU and 4GB RAM.

\textbf{\textit{Config-3}}: Qualcomm Snapdragon 855+ processor, Adreno 640 GPU and 8GB RAM.

The corresponding runtimes for processing images of half resolution of HD images on the devices can be found in Table \ref{mobile_runtime}. In comparison with other approaches, PyNet have the closest perceptual quality of our stacked DMSHN model, but it failed with an Out of Memory (OOM) error in these mid-range smartphones because of high memory consumption and instance normalization layers present in PyNet which are still not supported adequately by the TensorFlow Lite. However, our model is devoid of these problems.

\begin{table}[h]
\centering
\small
\begin{tabular}{|c|c|c|c|}
\hline
              Method & Config-1 & Config-2 & Config-3 \\ \hline
DMSHN       &   4.80       &      2.37   &    0.75     \\ \hline
Stacked DMSHN &     15.31     &  12.50       &    5.27     \\ \hline
\end{tabular}
\caption{Runtimes (in seconds) of our models on different smartphone configurations.}
\label{mobile_runtime}
\end{table}
\section{Conclusion}
In  this  paper, we  devised two end-to-end deep multi-scale networks, namely DMSHN and Stacked DMSHN for realistic bokeh effect rendering. Our  models  do  not  depend on any precomputed depth estimation maps or saliency maps and also do not require any other additional hardware (e.g. depth sensor or stereo camera) other than a monocular camera to aid the bokeh effect rendering. We see that Stacked DMSHN performs better than DMSHN both qualitatively and quantitatively. The Stacked DMSHN yields results of similar perceptual quality as PyNet \cite{pynet} and performs better than other approaches in the literature. Along with that, our proposed methods are lightweight, efficient, runnable in real time and also much faster than other competing approaches with good perceptual quality. We also showed that our models are deployable in mid-range smartphones too and take significantly less time where our closest competitor, PyNet \cite{pynet} is only able to run in high-end smartphones. In future, incorporation of dense connections in encoder and decoders and spatial attention can be explored to further improve the perceptual quality of rendered bokeh images.

{\small
\bibliographystyle{ieee_fullname}
\bibliography{cvpr}
}

\end{document}